\def\year{2021}\relax
\pdfoutput=1 
\documentclass[letterpaper]{article} % DO NOT CHANGE THIS
\usepackage{aaai21}  % DO NOT CHANGE THIS
\usepackage{times}  % DO NOT CHANGE THIS
\usepackage{helvet} % DO NOT CHANGE THIS
\usepackage{courier}  % DO NOT CHANGE THIS
\usepackage[hyphens]{url}  % DO NOT CHANGE THIS
\usepackage{graphicx} % DO NOT CHANGE THIS
\usepackage{natbib}  % DO NOT CHANGE THIS AND DO NOT ADD ANY OPTIONS TO IT
\usepackage{caption} % DO NOT CHANGE THIS AND DO NOT ADD ANY OPTIONS TO IT
\usepackage{soul}
\usepackage{booktabs,caption}
\usepackage[flushleft]{threeparttable}
\title{Improving Classification through Weak Supervision in Context-specific Conversational Agent Development for Teacher Education}
\author{

Debajyoti Datta \textsuperscript{\rm 1}, Maria Phillips \textsuperscript{\rm 1}, Jennifer Chiu \textsuperscript{\rm 2}, Ginger S. Watson \textsuperscript{\rm 2}, James P. Bywater \textsuperscript{\rm 3}, Laura Barnes \textsuperscript{\rm 1} and Donald Brown \textsuperscript{\rm 1}
    \\
}
\affiliations{
    \textsuperscript{\rm 1}Systems Engineering Department, University of Virginia\\
    \textsuperscript{\rm 2}Curry School of Education, University of Virginia\\
    \textsuperscript{\rm 3}James Madison University\\
    dd3ar@virginia.edu

}
\begin{document}

\maketitle
%Title Ideas
%Get rid of iterative development and IQA
% Focus is Conversational Agent to Improve Education Quality for Teachers in Training
%Methods: Weak Supervision, Active Learning and Transfer Learning
%Outcome: Generate Data more Efficiently, Classify More Accurately
%Application: Improved Assessment and Feedback tool, Transfer Learning

\begin{abstract}
Machine learning techniques applied to the Natural Language Processing (NLP) component of conversational agent development show promising results for improved accuracy and quality of feedback that a conversational agent can provide. The effort required to develop an educational scenario specific conversational agent is time consuming as it requires domain experts to label and annotate noisy data sources such as classroom videos. Previous approaches to modeling annotations have relied on labeling thousands of examples and calculating inter-annotator agreement and majority votes in order to model the necessary scenarios. This method, while proven successful, ignores individual annotator strengths in labeling a data point and under-utilizes examples that do not have a majority vote for labeling. We propose using a multi-task weak supervision method combined with active learning to address these concerns. This approach requires less labeling than traditional methods and shows significant improvements in precision, efficiency, and time-requirements than the majority vote method \cite{ratner2019training}. We demonstrate the validity of this method on the Google Jigsaw data set and then propose a scenario to apply this method using the Instructional Quality Assessment(IQA) to define the categories for labeling. We propose using probabilistic modeling of annotator labeling to generate active learning examples to further label the data. Active learning is able to iteratively improve the training performance and accuracy of the original classification model. This approach combines state-of-the art labeling techniques of weak supervision and active learning to optimize results in the educational domain and could be further used to lessen the data requirements for expanded scenarios within the education domain through transfer learning. 

\end{abstract}

\section{Introduction}

Research demonstrates that a key element of teacher education is giving preservice teachers the ability to rehearse high-leverage or evidence-based teaching practices \cite{grossman2009redefining} Training and practice of high-impact teaching practices have been shown to increase teacher effectiveness, confidence, and retention while increasing student achievement \cite{kilgo2015link}. However, development of these skills requires extensive practice and feedback. Teachers need more intentional, focused, and effective opportunities to build essential skills before and after they enter the classroom, yet for a variety of reasons these kinds of experiences are limited for most teacher education programs. 

Virtual human based simulations \cite{datta2016deep} have the potential to provide meaningful, deliberate practice for learning a wide range of teaching skills during pre-service teacher training and extended practice for advanced skills during in-service. Specifically, data-driven machine learning models have the potential to support rich, content-specific, classroom interactions and dialogue necessary for realistic and robust interactive classroom simulation in teacher training. These data-driven models coupled with the advancement in natural language processing offers the potential for meaningful teacher-to-virtual-student interactions that are scalable across a variety of technology platforms and across a teacher-education curriculum. With such a system, pedagogical moves and practices, such as teacher questioning, are automatically categorized and can be given back to the user to support reflection upon performance as well as to provide teacher educators with important data to guide their instruction. 

However, the development of these kinds of conversational agents are time consuming, require domain expertise as well as hundreds or thousands of hours to generate data, develop the conversational agent interface and to develop the Natural Language Processing (NLP) interface. A foundational element of a conversational agent is the ability to classify input from the user, this classification can be used as part of the response generation as well as a metric of assessment to provide feedback.

In order to develop a model to accurately classify inputs, a significant amount of labeled data using the assessment of choice is required. These labels would need to be generated by experts with domain knowledge and as of now it is not only difficult to find any such data set but it is also very time consuming to create. Based on the data generated for this research effort, it took labelers on average between 20 seconds to over 2 minutes to label a single data point depending on the labeler. While the exact amount of data needed will depend on many factors (error tolerance, quality of data, model complexity), a reference survey of the amount of training data required for different conversational agents shows between 3,000 data points to over 2 million \cite{smith_how_2019}. If the average time to label a point remains at 20 seconds then that results in 1,000 domain expert hours required just to generate a single scenario data set for a conversational agent. 

Artificial intelligence methods such as weak supervision, active learning, and transfer learning can significantly decrease the amount of labeled data points, thereby lessening the resources required to develop an accurate categorization mechanism. This mechanism, foundational to the quality of the assessment and feedback of a conversational agent, could then be improved and leveraged across a variety of scenarios in the education domain to further lessen the manual labels required through transfer learning. Weak supervision and additional methods discussed in this paper show promising results to not only these efficiencies in time and resources but also in improving the accuracies of predicted labeling leading to a more robust and realistic feedback and assessment tool. Improvements in labeling are a foundational element of most if not all Natural Language Processing tasks which could also be applied to other areas within the Education domain such as development of educational tools for children or conversational agents that act as tutors or provide teaching assistance. The goal of this research is proposing an improved method to support teacher-to-virtual-student dialogue using a conversational agent. We combine transfer learning, active learning, and weak supervision to bring efficiencies and improved outcome within this aspect of the education domain.  The work focused specifically on the effectiveness of these methods in developing a context-specific classroom training scenario in mathematics education.

%Since different approaches like transfer learning, active learning and weak supervision have different strengths in reducing the amount of labeled data required we combine them in different stages of the data collection pipeline and assess the efficiencies.
%The goal of the research was to assess the effectiveness of weak supervision to support teacher-to-virtual-student dialogue using a conversational agent. The work focused specifically on the effectiveness of weak supervision in developing a context-specific classroom training scenario in mathematics education.

The research discussed in this paper centers on the improvement of classification used in this initial data generation step as well as the classifier implemented in the conversational agent itself. Several questions this research addresses are:
\begin{enumerate}
  \item How can weak supervision be used effectively to develop data strategies in the education domain? 
  \begin{enumerate}
  \item How much data is required and how many annotators are needed for the development of a conversational agent for a specific task?
  \item What is the scalability of applying weak supervision methods for labeling data?
  \item What is the value of additional examples labeled by annotators?
  \end{enumerate}
  \item  How can domain knowledge be leveraged effectively to transfer knowledge to another part of the education domain?
\end{enumerate}

\section{Related Work}

\subsection{Conversational Agents in Education}
Conversational agents and assistants have been used for decades to facilitate learning. There are many examples of conversational agents used for educational and training purposes in K-12 and higher education, healthcare, military, and private industry settings. The most common forms of conversational agents in education are teaching agents who directly teach and support learning, peer agents who serve as knowledgeable learning companions to guide learners in the learning process, and teachable agents who function as a novice or less-knowledgeable student guided and taught by the learner who learns by teaching \cite{chhibber2019using}. Teaching and peer agents provide content, hints, prompts, advice, and feedback to the learner to inform, reinforce, correct, and scaffold the learner’s knowledge and skills that may be accomplished through a range of simplistic methods such as systematic prompts, decision trees, or more robust intelligent methods. Conversely, teachable agents must receive and interpret content, prompts, questions, and feedback. Teachable agents require AI-based approaches to understand, integrate, and assimilate information that the learner provides within and across lessons, units, and curricula. Research indicates that few conversational agents in education use AI but instead rely on simple decision trees \cite{smutny2020chatbots}. This reinforces the need for more research and development on the use of AI-based methods to support content-specific conversations with learners.

The current research addresses this need by investigating the effectiveness of weak supervision to reduce the need for hours of human-labeled data to facilitate natural language dialogue in the development of a teachable agent to facilitate teacher-to-virtual-student teaching dialogue. 

\subsection{Education Assessment Metric}
Many assessment tools have been developed for use in education, one such tool is the Instructional Quality Assessment (IQA). This assessment is a tool developed by the Learning Research and Developmental Center at the University of Pittsburgh since 2002 \cite{matsumura2006overview}.  The categories promoted in IQA have been validated by the initial developers of the method as well as additional research since it's creation \cite{boston2018instructional}. More recent developments suggest using IQA more than as an assessment tool but also as a feedback tool to help teachers improve their methods \cite{boston2018instructional}. 

The questions that teachers ask are essential for promoting students’ meaningful mathematical discourse. For example, the Academic Rigor component of the Instructional Quality Assessment \cite{junker2005overview, boston2012assessing}, which builds on earlier classifications of teacher questions (e.g. \cite{boaler2004importance}, distinguishes between probing or exploration questions that ask students to clarify their ideas or the connections between them, and procedural or factual questions that elicit facts or yes/no responses. The IQA instrument is intended to be used in contexts where cognitively demanding mathematical tasks are implemented, and is well suited for fine-grained teacher professional development such as that which focuses on teacher questioning \cite{boston2015comparison}.

%Consider changing to: 
%The assessment metric used for this study is the Instructional Quality Assessment (IQA; Boston, 2012). This assessment is a tool developed by the Learning Research and Developmental Center at the University of Pittsburgh since 2002 (Brian Junker and Yanna Weisberg 2006 [Citation]).  The categories promoted in IQA have been validated by the initial developers of the method as well as additional research since it's creation (Boston Amber 2018[Citation]). Building on earlier classifications of teacher questions (e.g. Boaler & Brodie, 2004), the IQA distinguishes between probing or exploration questions that ask students to clarify their ideas or the connections between them, and procedural or factual questions that elicit facts or yes/no responses. The instrument is intended to be used in contexts where cognitively demanding mathematical tasks are implemented, and is well suited for fine-grained teacher professional development such as that which focuses on teacher questioning (Boston, Bostic, Lesseig, & Sherman, 2015). The  IQA is suitable both as an assessment tool but also as a feedback tool to help teachers improve their methods (Boston Amber 2018[Citation]). 

% Conversational Agents are used in a wide variety of scenarios like healthcare \cite{datta2016deep}, education

\subsection{Data Collection}
The recent advances in machine learning in speech, text, and images mainly stem from large datasets. These datasets are often curated either by recording live data streams and then manually annotated in case of videos or collecting large text corpora and then manually annotated. There are dedicated platforms to do such annotations like Amazon Mechanical Turk, ScaleAI, Labelbox, etc. These platforms enable a wide range of support for creating a labeled dataset. For instance, the Labelbox platform enables both model assisted labeling and real-time human in the loop labeling (Active learning). In model-assisted labeling, annotators can quickly rely on annotations already generated by the machine. This approach has two benefits. One, it speeds up the annotation process \cite{masubuchi2020deep}, and secondly, human annotators can correct the incorrect predictions of the model.   

In any Machine Learning problem, data availability is a key to success and often obtaining enough data is a bottleneck in the process. In fact, it has been shown that often collecting larger samples of data addresses the issues of complex models. While significant amount of research in Machine Learning still centers around different model architectures effective data labeling strategies and data collection approaches can overcome significant huddles.

\subsection{Weak Supervision}
As we increase the amount of training data collected for a task, the accuracy of the model improves monotonically \cite{wang2020pretrain}. Collecting datasets at the scale of tasks like Semantic Entailment \cite{bowman2015large} is extremely challenging for new tasks like establishing a classifier for categories in IQA. Multiple studies have shown that when more data is collected, the model's accuracy increases without almost any change in the modeling architecture \cite{wang2020pretrain}. The authors show that simple LSTM models have performed within 1\% of the state of art accuracy models like BERT \cite{devlin2018bert} when the data set size has been continuously increased. So if the data pipeline is a critical component, then it is essential to figure out improved ways of labeling data and adding more signals to machine learning models.

%\subsection{Snorkel}
%[Reference paper from snorkel that proves the concept that we’re applying a method that works]
%“20.2 points in accuracy over a traditional supervised approach, 6.8 points over a majority vote baseline, and 4.1 points over a previously proposed weak supervision method that models tasks separately” [Cite] - 
%[What hasn’t been done is it hasn’t been scaled to show that it works in all types of data sets, this is something that we show - and then apply to education]
%[Equations]
Weak supervision  is a machine learning paradigm that trains models by incorporating data labeled or generated through machine learning models and crowdsourced workers effectively \cite{dawid1979maximum, ratner2016data, karger2011iterative, bunescu2007learning}. Noisy labels used in weak supervision are often cost-effective to acquire. Weak supervision has been applied successfully for knowledge based completion \cite{hoffmann2011knowledge}, sentiment analysis \cite{severyn2015twitter} and in the biomedical domain \cite{wallace2016extracting, sorin2020deep}. The scalability of this approach with regards to the number of annotators and number of examples required for weak supervision strategies to be effective has only been demonstrated in a theoretical context. Our research demonstrates the validity using an external dataset, Google Jigsaw Toxicity dataset \cite{wulczyn2017ex}. We also do ablation studies for the number of examples required and the number of annotators required for a given accuracy.

\subsection{Active Learning}
%Define what they are/state of the art
%(What is the mathematical derivation of)
%Foundational reference citation 
%Here’s how it’s been to applied to education

In active learning  the goal is to reduce the number of labeled examples required by querying the users for labels of the most uncertain examples or the most informative examples \cite{settles2009active}. In this setting, from a limited number of labeled examples, and a pool of unlabeled examples different algorithmic approaches can be adapted to select examples for labeling. These examples can be to increase the diversity of examples, (also known as diversity sampling) or examples about which the model is least confident (uncertainty sampling). In our scenario, we select examples for active learning by randomly selecting from one of the strategies below:
\begin{itemize}
  \item Examples that have most number of conflicting labels from our rule based labeling functions.
  \item Examples that have the least number of labels from labeling functions. For instance if most labeling functions have abstained from labeling a data point then we select the least labeled point.
\end{itemize}

% \subsection{Transfer Learning}
% %Define what they are/state of the art
% %(What is the mathematical derivation of)
% %Foundational reference citation 
% %Here’s how it’s been to applied to education

\subsection{Transfer learning in NLP}
Developments in the modeling approach in Natural Language Processing (NLP) have centered around popular benchmarks like Glue \cite{wang2018glue} and SuperGlue \cite{wang2019superglue} tasks that evaluate the performance of a model across a whole range of NLP tasks. These tasks are generic like Semantic Entailment \cite{bowman2015large}, or CoLA \cite{warstadt2018neural} and require large labeled datasets that are often collected through crowdsourced platforms. This approach helps create standard benchmarks and evaluation protocols, but Natural Language Benchmarks do not necessarily evaluate the task properly \cite{gururangan_annotation_2018}. Solutions like \cite{gardner_evaluating_2020} and \cite{kaushik_learning_2020} have been proposed but both lack sufficient theoretical grounding as was shown by (Anonymous et al., 20XX). \cite{nie2019adversarial} seem to address this but are still limited to specific tasks and datasets. Creating datasets for new tasks, like those using the IQA, remains a challenge especially for tasks requiring the collection and labeling of larger datasets. 

Another approach to improve the efficiencies in conversational agent development is transfer learning. In transfer learning, a model is trained on a generic task like unsupervised language modeling and subsequently fine-tuned on a downstream task like semantic entailment \cite{bowman2015large}. Pre-training data sets for images involves training on the large-scale Imagenet data sets \cite{russakovsky2015imagenet} with thousands of labels, but pre-training is unsupervised for language. Fine-tuning on downstream tasks \cite{devlin2018bert}, \cite{howard2018universal} still requires task specific labeled datasets and the quality of the downstream tasks still depend on the amount of labels and examples available. The need for labeled datasets is still a bottleneck, and higher quality labeled datasets determine a task's success and a benchmark. 

One alternative to fine-tuning is to rely on features generated from BERT and directly use the resulting model for downstream tasks \cite{wang2020pretrain}. This approach is scalable and significantly faster than fine-tuning an entire network.  

\section{Methodology}

\begin{figure*}[ht]
\centering
\includegraphics[width=0.8\textwidth]{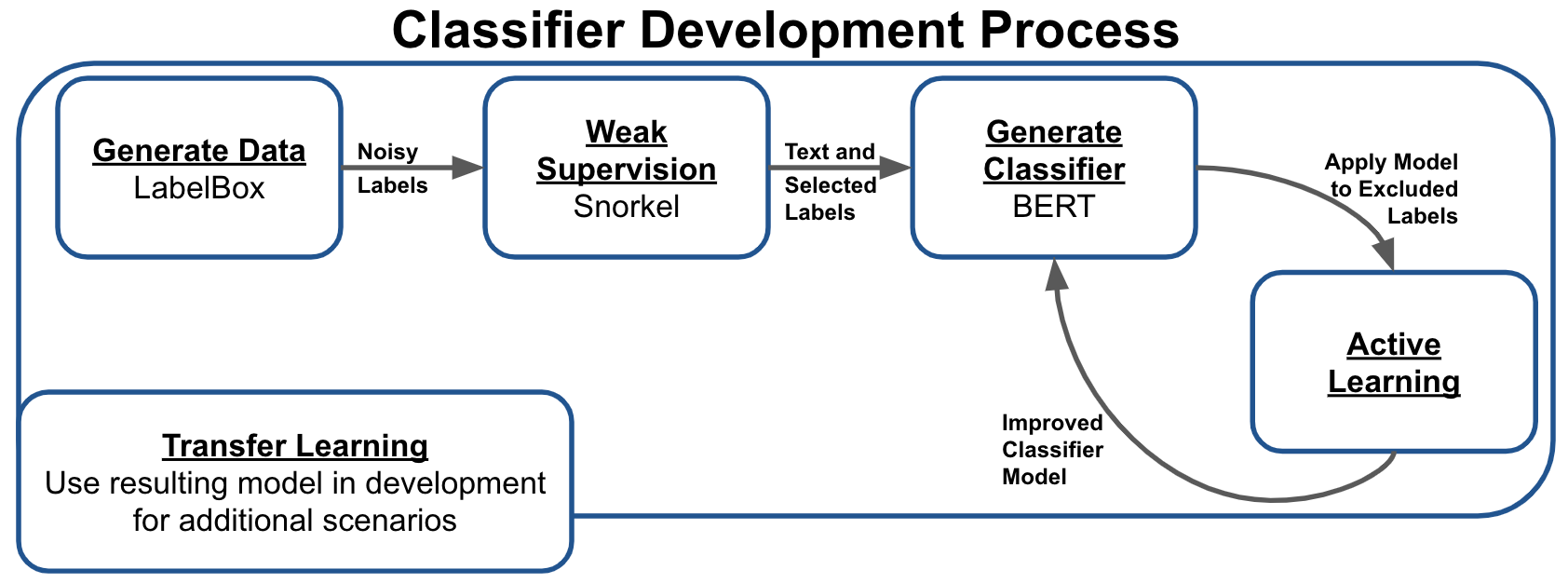} % Reduce the figure size so that it is slightly narrower than the column.
\caption{This shows the approach used in improving the classifier portion of a conversational agent}
\label{fig10}
\end{figure*}

Our approach is to use aspects of weak learning to improve the efficiency and accuracy of data labeling. We then use these improved labels to train a classification model and combine this model with active learning methods to improve the resulting classification model. This process can then be used in transfer learning to be applied to the NLP portion for additional scenario development of conversational agents assuming the same assessment metric is desired. This process is depicted in Figure \ref{fig10}

%Added Here so it will be on methodology page

\subsection{Implementation in Education Domain}

\subsubsection{Assessment Metric}
IQA is the chosen assessment rubric that is used to create the categories to be predicted by the model. The categories guidelines are expanded on below. 
\begin{itemize}
    \item \textbf{Probing and Exploring}: Clarifies student thinking, enables students to elaborate their own thinking for their own benefit and the class. Points to underlying mathematical relationships and meanings and makes links among mathematical ideas. (e.g. Explain to me how you got that expression?)

    \item \textbf{Procedural or Factual}: Elicits a mathematical fact or procedure; Requires a yes/no or single response answer; Requires the recall of a memorized fact or procedure (e.g. What is the square root of 4?)
    \item \textbf{Other Mathematical}: Related to teaching and learning mathematics but do not require mathematical procedures or factual knowledge; probe students’ thinking, press for explanations. (eg. Which problem was the most difficult?)
    \item \textbf{Non-Mathematical}: Does not relate to teaching and learning mathematics (e.g. Why didn’t you use graph paper?)
\end{itemize}

Of note is while initial labeling occurred, an additional "Expository" category to account for rhetorical utterances and instructional statements in question form. This category was excluded from analysis for the results published in this paper.

\subsubsection{Data}
The education data was collected by audio recording the teacher-student conversations that took place in a high-school Geometry session while students were engaging with a project aimed at helping them understand the relationship of scale factor on the dimensions, surface area, and volume of rectangular prisms. %[I don't think we need this but maybe we should ask? ]Four teachers (all white, two female and two male, average teaching experience approximately six years) wore lapel mics during a total of seven classes, interacting with 68 students. The school was located in a small city in the rural mid-Atlantic region of the United States, and the student demographics of the school were 12\% Black, 44\% Hispanic, and 38\% White students, with 68\% of the students receiving free or reduced lunch and 39\% of the students classified as having Limited English Proficiency.
 
The project involved students interacting with visualizations of a rectangular prism, using a data table to record observations from the visualization, and explaining the impact of scale factor. The teacher-student conversations related to the explanations that students were writing or revising.

Audio recordings were trimmed to only include periods of the class where one-to-one teacher-student conversations about the impact of scale factor occurred. The 270 minutes of trimmed classroom audio was then transcribed using the Rev.com translation service. The transcript was reviewed by a member of the research team, and transcription errors were corrected, names of students and teachers replaced with personal pronouns to anonymize the transcript, and sections of text that did not related to the project (such as conversations about using the bathroom, tardiness, laptop charging, room temperature) were removed.

Using these transcripts, each utterance or sentence spoken by the teacher was separated as a data point to be labeled. This method resulted in 867 sentences that are spoken by the teacher in the conversations. We took a random subset of twenty percent of these (174) to be labeled and this is the data used for our initial experiments. 

Three certified educators with backgrounds in teacher education were trained to rate instances of teacher dialogue into one of the four categories defined for the IQA. Before the rating of the data set, raters met and reviewed the criteria for each level of the IQA with an associated range of examples. Then raters labeled a common evaluation set of approximately 50 teacher questions. After all rating of the common evaluation set was complete, raters met to review and resolve variations in ratings and interpretations of IQA categories before moving onto the the larger data set. 

%I think this needs to move to dif section?

%If we talk about future desires I think it should go in the conclusion/future works section.
%Our eventual goal is to evaluate and have a continuous feedback system for trainees such that they can continuously improve the teaching methodologies and approaches. focus centers around effective data labeling strategies and development methodologies for generating new educational scenarios and new contexts of training scenarios. 
%I don't htink this is IQA specific, it's an educational assessment tool/ don't think we have room to go into it
%The nature of IQA is such that some things like probing questions can be transferred between teaching fractions versus teaching algebra but understanding mathematical contexts of a task is distinctly different between algebra and fractions. 

In order to address domain specific tasks we use transfer learning strategies. In fact repeated iterations are often necessary to come up with clearer rubric for what constitutes different categories of questions and their corresponding descriptions. 

\subsubsection{Weak Supervision and Active Learning}

For the initial scenario development, we relied on expert labels in a visual platform that enabled the annotators to select a specific category for the question. For the first iteration of the data collection, the  annotation coverage over the data set was nearly 100\%. In other words, all data points were labeled by all the annotators. The small number of data points serves multiple purposes: Revising annotation guidelines and selecting the complexity of tasks and labels for specific subcategories. We randomly selected out the 250 examples and specifically computed the accuracy for each of the subcategories of tasks. If the annotation agreement was low for one subtask, like "probing questions", we could specifically query for annotators to label more probing questions than any other category. This process can be iteratively run as many times until the desired accuracy is obtained. Simple ML models that rely on very little labeled data can be used for model-assisted labeling to improve data quality. In model assisted labeling a machine learning model already provides the prediction and the annotator either accepts the prediciton or changes it. It has been shown that model assisted labeling speeds up annotation time required per annotation \cite{masubuchi2020deep}, and significant disagreements between annotators and models found through model assisted labeling are good candidates for active learning. 

\begin{figure*}[ht]
\centering
\includegraphics[width=0.8\textwidth]{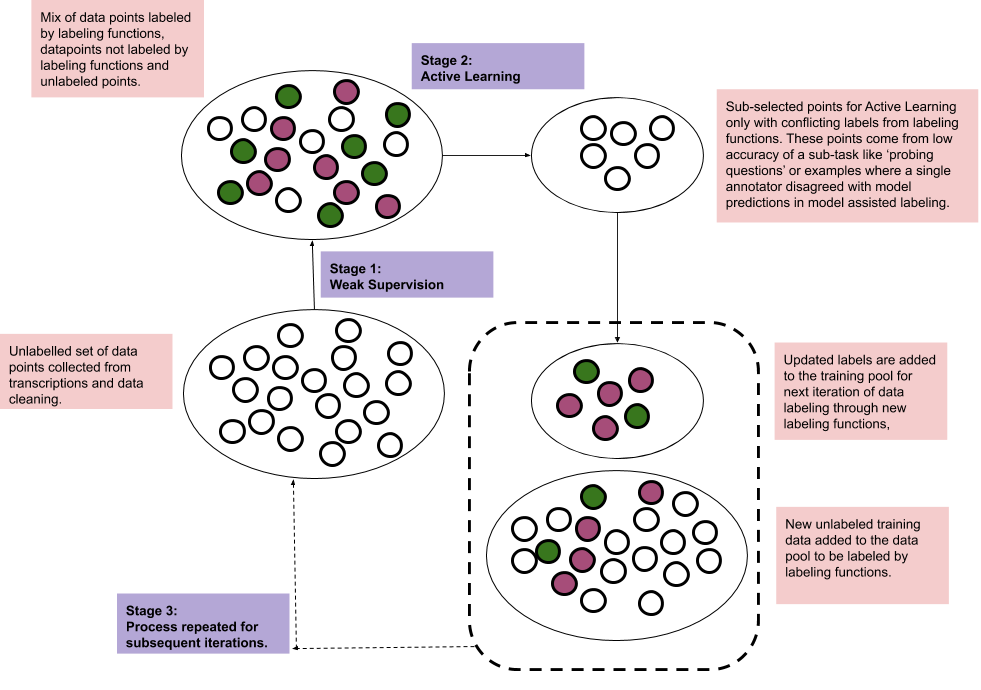} % Reduce the figure size so that it is slightly narrower than the column.
\caption{The sampling strategy to label data at each stage is selected through a combination of task specific metrics and examples which had conflicting outputs from model assisted labeling. Since data is also programmatically labeled with updated labeling functions in each stage, this process is significantly faster than traditional active learning strategy. This also gives an opportunity to revise rule based labeling functions. If the coverage of the new labeling function is low or the empirical accuracy in the validation set is low, the labeling function is discarded and data labeled by the labeling function is added back to the pool.}
\label{fig5}
\end{figure*}
\subsubsection{Model Description and Labeling functions}

The text sentences were converted to features using the state of the art BERT model's feature outputs for a sentence. We did not fine-tune the model, but instead used these as input features and performed a logistic regression. The labels for the generative models were used to train the logistic regression model. We performed five sets of experiments for hyperparameter tuning of the logistic regression model for this education scenario, and the best model was chosen based on validation set accuracy. This validation set label was generated through noisy labels of the weak supervision task. 

The amount of data within the education scenario is limited. In order to illustrate the validity and scalability of our approach with the number of examples and number of annotators we relied on the Google toxicity dataset \cite{wulczyn2017ex} and used weak supervision approaches on it. We were able to illustrate the the strengths of the discussed weak supervision methods in terms of the number of annotators and the number of examples labeled by them.

\subsection{Quantifying relationship of annotators and number of examples}

%NOTE: NEED to add motivation for why we are including toxicity dataset - basically the data we got for education is minimal an we have few annotators, in order to help validate the methods we suggest we wanted to show that these methods not only scale well but also provide an opportunity to assess an optimal strategy (how many annotators, how much data should they label etc)

\begin{figure}[t]
\centering
\includegraphics[width=1\columnwidth]{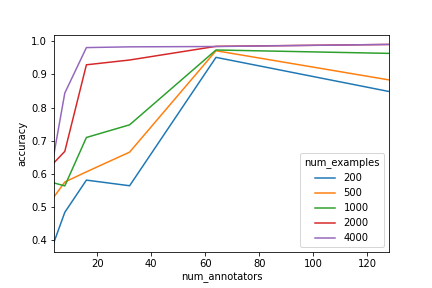} % Reduce the figure size so that it is slightly narrower than the column. Don't use precise values for figure width.This setup will avoid overfull boxes.
\caption{The number of annotators vs the number of examples each annotator is annotating. }.
\label{fig1}
\end{figure}

\begin{figure}[t]
\centering
\includegraphics[width=1\columnwidth]{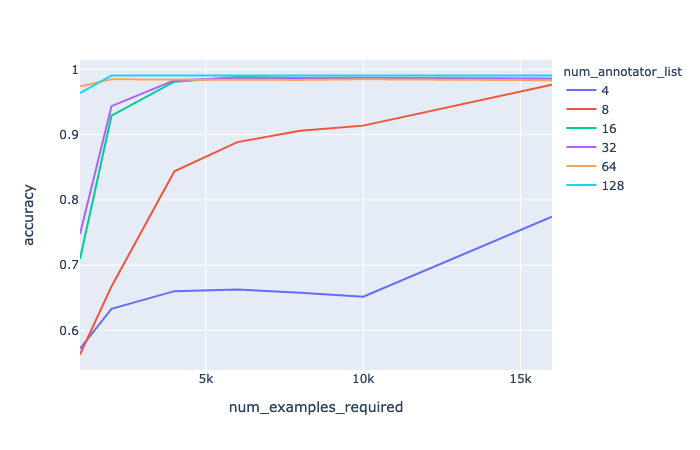} % Reduce the figure size so that it is slightly narrower than the column. Don't use precise values for figure width.This setup will avoid overfull boxes.
\caption{Relationship of Accuracy and the minimum number of examples required given the number of annotator list.}.
\label{fig2}
\end{figure}

The toxicity dataset \cite{wulczyn2017ex} is a large scale dataset of 100k high quality human labeled comments.  Using the noisy labels generated in the toxicity data set, we use a method developed by  \cite{ratner2016data} which explicitly models this as a generative process by learning the accuracy of the labeling functions and the corresponding dependencies between them. A  discriminative model that relies on the labels generated through this process is trained for downstream tasks like classification. One approach to developing labeling functions is to rely on rules like \cite{neeraj2020wsfortoxicity} but in our experiments we only model annotators as labeling functions. The toxicity challenge has a highly unbalanced dataset, with only 10\% of the comments as toxic. In order to address this, we created stratified sampling for the test set. Since the test set was large, we selected a more balanced distribution of positive and negative classes for the evaluation. For fairness, the above accuracy numbers in the two Figures \ref{fig1} and  \ref{fig2}, exactly half toxic and half non-toxic comments, were selected. This prevents a skewed baseline and helps conduct ablation studies in measuring the actual performance of annotators and the number of examples. The experiment set up is as follows:

%I don't think we need this senetence: In large datasets with many annotators, high coverage is not critical; higher coverage helps when the amount of data and number of annotators is small. 

\noindent\textbf{Number of annotators selected based on high coverage} In Figure \ref{fig1} the toxicity dataset, we scale up the number of annotators progressively and the number of examples they are annotating. So we select the top N annotators with the highest number of examples annotated, representing the x-axis. Since this data has already been collected, we simulate the experiments by the most annotated examples from the dataset. This increases the coverage and increases the likelihood of selecting examples that were annotated by the annotators. Given high coverage, it is clear that as we scale up the number of annotators, the model's accuracy increases. Keeping the number of annotators constant, if annotators label more examples each, the accuracy increases. The slight dip for 200 and 500 examples each above 50 annotators is likely because the coverage drops sharply when we increase the number of annotators. However, the performance consistently improves when we increase the number of examples annotated by each annotator, and there is no significant gain between annotating 1000, 2000, and 4000 examples when we increase the number of annotators. This approach can lead to  significant cost savings by either increasing the coverage through several examples annotated without increasing the number of annotators for tasks.

\noindent\textbf{Number of examples annotated vs number of annotators} In the second ablation study, we measure the performance of the minimum number of examples required vs. accuracy. Figure \ref{fig2} the experiment set up is as follows. Given a certain number of annotators we measure the minimum number of examples required for annotation to achieve different accuracy levels. The baseline is 0.5 since it is a binary classification task. The accuracy achieved increases as we increase the number of annotators make a significant difference up to 8 annotators. But with 16 annotators, the gain is minimal by scaling the number of examples highlighting the importance of the annotators' coverage of the labeling data. Note that with 16 annotators, labeling 1300 examples gives superior performance compared to 8 annotators labeling ten times that many examples. The number of annotators makes a much more significant difference than just scaling the number of annotated examples. Also, notice the minimal gain achieved by increasing both the number of annotators and the number of examples beyond a certain point. (16 annotators labeling 4000 examples each). In this dataset, there is no benefit to increasing the number of annotators or examples to achieve accuracy higher than 95\%. 

\section{Results}

Data labeling through weak supervision using the IQA defined categories has enabled us to collect data more rapidly than traditional methods. The inter-annotator agreement for the initial labeling tasks was 49.4 (Cohen's Kappa), and it subsequently improved to 79.4 in the second iteration of data labeling. The labeling functions were designed using multi-task methods of weak supervision. These functions were then used to generate a labeled data set that outperformed majority vote labeling and were used to create a classification model using BERT. Active learning  was then used with this classification model to label examples that did not have a majority vote in either iteration. The test set for evaluation of the model were examples collected through the second round of data collection. This was an optimal way to improve the annotators time since the annotators in our case were expert educators. Since a new category was added in the subsequent iteration of data collection, we report individual class level accuracy for examples that had at least 50 samples per category.  

Labeling functions were iteratively developed by using coverage and accuracy on the test set as an attribute. Since conversation from videos often had 'pauses' and incomplete sentences, iterative labeling function development was necessary to increase the accuracy on the test set. Our approach efficiently combines multiple methods (weak supervision, active learning and transfer learning) to address the goal of developing educational training scenarios quickly as well as developing an improved classification and thereby feedback tool that can be used in a conversational agent within the education domain.

\begin{table}[ht]
\caption{Class Level Accuracy}
\centering
\begin{tabular}{ll}
\ul \textbf{Category} & {\ul \textbf{Accuracy}} \\
              Non-math &  0.742               \\
              Procedural  & 0.815                \\
              Probing or Exploring  & 0.695             
\end{tabular}
\end{table}

%\begin{table}[1]
%\centering
%\resizebox{.95\columnwidth}{!}{
%\begin{tabular}{l|l}
 %   non-math & 0.742 \\
  %      procedural & 0.815 \\
 %probing\_exploring & 0.695 \\
%\end{tabular}
%\caption{LaTeX style packages that must not be used.}
%\label{table2}
%\end{table}

\begin{table}
\caption{Labeling Function Results}
\begin{tabular}{rrrrr}

 \textbf{ID}  &  \textbf{Overlap} &  \textbf{Conflict} &  \textbf{Correct} &  \textbf{Accuracy} \\

    LF1 &     0.988 &      0.468 &      184 &   0.749 \\
    LF2 &     1.000 &      0.472 &      199 &   0.796 \\
    LF3 &     0.980 &      0.468 &      201 &   0.820 \\

\end{tabular}
\begin{tablenotes}
      \small
      \item *LF: Label Function
    \end{tablenotes}
\end{table}

By relying on active learning using the conflict strategy (where multiple annotators disagreed), our model was able to achieve an accuracy of 62.85\% on examples that had a conflict. The individual class level accuracy on the test set has been reported in Table 1.

\section{Conclusion}

In this paper we measured the effectiveness of weak supervision in developing an education based training scenario using the IQA as a metric. In order to account for the amount of data required we used active learning along with weak supervision to improve on the conflicting labels. Since the framework described here is generic we believe this can be applied to any training scenario. We also did an ablation study of the impact of the number of workers and number of examples with accuracy on a large publicly available dataset to highlight the empirical evidence at scale. This approach will be used in future studies in developing other mathematical scenarios like Fractions and Algebra. Effective usage of noisy annotators will also help balance the cost of training annotators and using noisy labels. The above described labeling functions also enables us to use transfer learning for some categories and rely in domain specific labeling functions for others. We will also open source our entire code and implementation on the publicly available dataset for replicating the key graphs.

%[Include the impact and future efforts and considerations for these results (shortcomings of the study)]
%What does the efficiency mean for the virtual education field?
%[Time required to do these tasks- for labeling, for hands on training resources, other motivation for education domain value added - amount of time required to label the rest of the data]

% \subsection{Future Efforts}
%[-Include that we have not yet validated the dataset - what are ways to validate and get a ground truth? If snorkel has shown this method is better than majority votes, is there a better way to pick a ground truth than majority votes?
%-Try to increase annotator numbers/data strategy to reflect what toxicity shows is a good strategy]

% \begin{table}[t]
% \centering
% %\resizebox{.95\columnwidth}{!}{
% \begin{tabular}{l|l|l|l}
%     authblk & babel & cjk & dvips \\
%     epsf & epsfig & euler & float \\
%     fullpage & geometry & graphics & hyperref \\
%     layout & linespread & lmodern & maltepaper \\
%     navigator & pdfcomment & pgfplots & psfig \\
%     pstricks & t1enc & titlesec & tocbind \\
%     ulem
% \end{tabular}
% \caption{LaTeX style packages that must not be used.}
% \label{table2}
% \end{table}

% Using the \centering command instead of \begin{center} ... \end{center} will save space
% Positioning your figure at the top of the page will save space and make the paper more readable
% Using 0.95\columnwidth in conjunction with the

\section{Acknowledgments}
Want to acknowledge Labelbox for providing us an academic license.
This work was supported in part by an Anonymous Grant Foundation. %Robertson
\bibliography{paper}

\end{document}